\documentclass[11pt,a4paper]{article}
\usepackage[hyperref]{acl2019}
\usepackage{times}
\usepackage{latexsym}

\usepackage{url}

\newcommand{\lineacross}{\rule{\linewidth}{1pt}}

\aclfinalcopy 

\title{Natural Language Generation Challenges for Explainable AI}

\author{Ehud Reiter \\
  University of Aberdeen \\
  \texttt{e.reiter@abdn.ac.uk} \\}

\date{}

\begin{document}
\maketitle
\begin{abstract}
  Good quality explanations of artificial intelligence (XAI) reasoning must be written (and evaluated) for an explanatory purpose, targeted towards their readers, have a good narrative and causal structure, and highlight where uncertainty and data quality affect the AI output.   I discuss these challenges from a Natural Language Generation (NLG) perspective, and highlight four specific ``NLG for XAI'' research challenges.
\end{abstract}

\section{Introduction}

Explainable AI (XAI) systems \cite{biran2017explanation,gilpin2018explaining} need to explain AI reasoning to human users.  If the explanations are presented using natural languages such as English, then it is important that they be accurate, useful, and easy to comprehend.   Ensuring this requires addressing challenges in Natural Language Generation (NLG) \cite{reiter2000building,gatt2018survey}.

Figure~\ref{ExampleExplanation} gives an example of a human-written explanation of the likelihood of water or gas being close to a proposed oil well; I chose this at random from many similar explanations in a Discovery Evaluation Report \cite{equinorDiscovery} produced for an oil company  which was deciding whether to drill a well.  Looking at this report, it is clear that
\begin{itemize}
    \item It is {\em written for a purpose} (helping the company decide whether to drill a well), and needs to evaluated with this purpose in mind.  For example, the presence of a small amount of water would not impact the drilling decision, and hence the explanation is not ``wrong'' if a small amount of water is present.
    \item It is {\em written for an audience}, in this case specialist engineers and geologists, by using specialist terminology which is appropriate for this group, and also by using vague expressions (e.g., ``minor amount'') whose meaning is understood by this audience.   A report written about oil wells for the general public (such as \citet{national2011deep}) uses very different phrasing.
    \item It has a {\em narrative structure}, where facts are linked with causal, argumentative, or other discourse relations.   It is not just a list of observations.
    \item It explicitly {\em communicates uncertainty}, using phrases such as ``possibility'' and ``unlikely'',
\end{itemize}

\begin{figure}
\lineacross{}
It is also unlikely that a water or gas contact is present very close to the well. During the DST test, the well produced only minor amounts of water.  No changes in the water content or in the GOR of the fluid were observed.  However, interpretation of the pressure data indicates pressure barriers approximately 65 and 250m away from the well [...]  It is therefore a possibility of a gas cap above the oil.  On the other hand, the presence of a gas cap seems unlikely due to the fact that the oil itself is undersaturated with respect to gas (bubble point pressure = 273 bar, reservoir pressure = 327.7 bar)
\caption{Example of a complex explanation}
\label{ExampleExplanation}
\lineacross{}
\end{figure}

If we want AI reasoning systems to be able to produce good explanations of complex reasoning, then these systems will also need to adapt explanations to be suitable for a specific purpose and user, have a narrative structure, and communicate uncertainty.  These are fundamental challenges in NLG.

\section{Purpose and Evaluation}
A core principle of NLG is that generated texts have a {\em communicative goal}. That is, they have a purpose such as helping users make decisions (perhaps the most common goal), encouraging users to change their behaviour, or entertaining users.    Evaluations of NLG systems are based on how well they achieve these goals, as well as the accuracy and fluency of generated texts.  Typically we either directly measure success in achieving the goal or we ask human subjects how effective they think the texts will be at achieving the goal \cite{gkatzia2015snapshot}.

Real-world explanations of AI systems similarly have purposes, which include
\begin{itemize}
    \item Helping developers {\em debug} their AI systems.  This is not a common goal in NLG, but seems to be one of the most common goals in Explainable AI.   The popular LIME model \cite{ribeiro2016should}, for example, is largely presented as a way of helping ML developers choose between models, and also improve models via feature engineering.
    \item Helping users detect mistakes in AI reasoning ({\em scrutability}).   This is especially important when the human user has access to additional information which is not available to the AI system, which may contradict the AI recommendation.  For example, a medical AI system which only looks at the medical record cannot visually observe the patient; such observations may reveal problems and symptoms which the AI is not aware of.
    \item Building {\em trust} in AI recommendations.  In medical and engineering contexts, AI systems usually make recommendations to doctors and engineers, and if these professionals accept the recommendations, they are liable (both legally and morally) if anything goes wrong.  Hence systems which are not trusted will not be used.
\end{itemize}
The above list is far from complete, for example \citet{Tintarev2012} also include Transparency, Effectiveness, Persuasiveness, Efficiency, and Satisfaction in their list of possible goals for explanations.

Hence, when we evaluate an explanation system, we need to do so in the context of its purpose.   As with NLG in general, we can evaluate explanations at different levels of rigour.   The most popular evaluation strategy in NLG is to show generated texts to human subjects and ask them to rate and comment on the texts in various ways.  This is leads to my first challenge

\begin{itemize}
    \item  {\em Evaluation Challenge}: Can we get reliable estimates of scrutabilty, trust (etc) by simply asking users to read explanations and estimate scrutability (etc)?  What experimental design (subjects, questions, etc) gives the best results?  Do we need to first check explanations for accuracy before doing the above?
\end{itemize}

Other challenges include creating good experimental designs for task-based evaluation, such as the study \citet{biran2017human} did to assess whether explanations improved financial decision making because of increased scrutability; and also exploring whether automatic metrics such as BLEU \cite{papineni2002bleu} give meaningful  insights about trust, scrutability, etc.

\section{Appropriate Explanation for Audience}
A fundamental principle of NLG is that texts are produced for users, and hence should use appropriate content, terminology, etc for the intended audience \cite{paris2015user,walker2004}.   For example, the Babytalk systems generated very different summaries from the same data for doctors \cite{portet2009automatic}, nurses \cite{hunter2012automatic}, and parents \cite{mahamood2011generating}.

Explanations should also present information in appropriate ways for their audience, using features, terminology, and content that make sense to the user \cite{lacave2002review,biran2017human}.   For example, a few years ago I helped some colleagues evaluate a system which generated explanations for an AI system which classified leaves \cite{alonso2017exploratory}.   We showed these explanations to a domain expert (Professor of Ecology at the University of Aberdeen), and he struggled to understand some explanations because the features used in these explanation were not the ones that he normally used to classify leaves.

Using appropriate terminology (etc) is probably less important if the goal of the explanation is debugging, and the user is the machine learning engineer who created the AI model.  In this case, the engineer will probably be very familiar with the features (etc) used by the model.   But if explanations are intended to support end users by increasing scrutability or trust, then they need to be aligned with the way that users communicate and think about the problem.

This relates to a number of NLG problems, and I would like to highlight the below as my second challenge:
\begin{itemize}

    \item {\em Vague Language Challenge:}  People naturally think in qualitative terms, so explanations will be easier to understand if they use vague terms \cite{van2012not} such as ``minor amount'' (in Figure~\ref{ExampleExplanation}) when possible.   What algorithms and models can we use to guide the usage of vague language in explanations, and in particular to avoid cases where the vague language is interpreted by the user in an unexpected way which decreases his understanding of the situation?
\end{itemize}
There are of course many other challenges in this space.  At the content level, it would really help if we could prioritise messages which are based on features and concepts which are familiar to the user.  And at the lexical level, we should try to select terminology and phrasing which make sense to the user.

\section{Narrative Structure}

People are better at understanding symbolic reasoning presented as a narrative than they are at understanding a list of numbers and probabilities \cite{kahneman2011thinking}.   ``John smokes, so he is at risk of lung cancer'' is easier for us to process than ``the model says that John has a 6\% chance of developing lung cancer within the next six years because he is a white male, has been smoking a pack a day for 50 years, is 67 years old, does not have a family history of lung cancer, is a high school graduate [etc]''.  But the latter of course is the way most computer algorithms and models work, including the one I used to calculate John's cancer risk\footnote{https://shouldiscreen.com/English/lung-cancer-risk-calculator}.   Indeed, \citet{kahneman2011thinking} points out that doctors have been reluctant to use regression models for diagnosis tasks, even if objectively the models worked well, because the type of reasoning used in these models (holistically integrating evidence from a large number of features) is not one they are cognitively comfortable with.

The above applies to information communicated linguistically.  In contexts that do not involve communication, people are in fact very good at some types of reasoning which involve holistically integrating many features, such as face recognition.  I can easily recognise my son, even in very noisy visual contexts, but I find it very hard to describe him in words in a way which lets other people identify him.

In any case, linguistic communication is most effective when it is structured as a narrative.  That is, not just a list of observations, but rather a selected set of key messages which are linked together by causal, argumentative, or other discourse relations.   For example, the most accurate way of explaining a smoking risk prediction based on regression or Bayesian models is to simply list the input data and the models result.
\begin{quote}
``John is a white male.  John has been smoking a pack a day for 50 years. John is 67 years old. John does not have a family history of lung cancer. John is a high school graduate.  John has a 6\% chance of developing lung cancer within the next 6 years.''
\end{quote}
But people will probably understand this explanation better if we add a narrative structure do it, perhaps by identifying elements which increase or decrease risks, and also focusing on a small number of key data elements \cite{biran2017human}.
\begin{quote}
``John has been smoking a pack a day for 50 years, so he may develop lung cancer even though he does not have a family history of lung cancer.''
\end{quote} 
This is not the most accurate way of describing how the model works (the model does not care whether each individual data element is ``good'' or ``bad''), but it probably is a better explanation for narrative-loving humans.

In short, creating narratives is an important challenge in NLG \cite{reiter2008importance}, and its probably even more important in explanations.  Which leads to my third challenge
\begin{itemize}
    \item {\em Narrative Challenge:}   How can we present the reasoning done by a numerical non-symbolic model, especially one which holistically combines many data elements (e.g., regression and Bayesian models) as a narrative, with key messages linked by causal or argumentative relations?
\end{itemize}

\section{Communicating Uncertainty and Data Quality}
People like to think in terms of black and white, yes or no; we are notoriously bad at dealing with probabilities \cite{kahneman2011thinking}.  One challenge which has received a lot of attention is communicating risk \cite{berry2004risk,lundgren2018risk}; despite all of this attention, it is still a struggle to get people to understand what a 13\% risk (for example) really means.   Which is a shame, because effective communication of risk in an explanation could really increase scrutability and trust.

Another factor which is important but has received less attention than risk is communicating data quality issues.    If we train an AI system on a data set, then any biases in the data set will be reflected in the system's output,   For example, if we train a model for predicting lung cancer risks purely on data from Americans, then that model may be substantially less accurate if it is used on people from very different cultures.  For instance, few Americans grow up malnourished or in hyper-polluted environments; hence a cancer-prediction model developed on Americans may not accurately estimate risks for a resident of Delhi (one of the most polluted city in the world) who has been malnourished most of her life.   Any explanation produced in such circumstances should highlight training bias and any other factors which reduce accuracy.

Similarly, models (regardless of how they are built) may produce inaccurate results if the input data is incomplete or incorrect.   For example, suppose someone does not know whether he has a family history of lung cancer (perhaps he is adopted, and has no contact with his birth parents).   A lot of AI models are designed to be robust in such cases and still produce an answer; however, their accuracy and reliability may be diminished.  In such cases, I think explanations which are scrutable and trustworthy need to highlight this fact, so the user can take this reduced accuracy into consideration when deciding what to do.

There has not been much previous research in data quality in NLG (one exception is \citet{inglis2017textually}), which is a shame, because data quality can impact many data-to-text applications, not just explanations.  But this does lead to my fourth challenge

\begin{itemize}
    \item {\em Communicating Data Quality Challenge:}   How can we communicate to users that the accuracy of an AI system is impacted either by the nature of its training data, or by incomplete or incorrect input data?
\end{itemize}
Of course, communicating uncertainty in the sense of probabilities and risks is also a challenge for both NLG in general and explanations specifically!

\section{Conclusion}
If we want to produce explanations of AI reasoning in English or other human languages, then we will do a better job if we address the key natural language generation issues of evaluation, user-appropriateness, narrative, and communication of uncertainty and data quality.   I have in this paper highlighted four specific challenges within this areas which I think are very important in generating good explanations:
\begin{itemize}
    \item {\em Evaluation:} Develop ``cheap but reliable'' ways of estimating scrutability, trust, etc.
    \item {\em Vague Language:} Develop good models for the use of vague language in explanations.
    \item {\em Narrative:} Develop algorithms for creating narrative explanations.
    \item {\em Data Quality:}  Develop techniques to let users know how results are influenced by data issues.
\end{itemize}
All of these are generic NLG challenges which are important across the board in NLG, not just in explainable AI.

\section*{Acknowledgments}
This paper started off as a (much shorter) blog \url{https://ehudreiter.com/2019/07/19/nlg-and-explainable-ai/}.  My thanks to the people who commented on this blog, as well as the anonymous reviewers, the members of the Aberdeen CLAN research group, the members of the Explaining the Outcomes of Complex Models project at Monash, and the members of the NL4XAI research project, all of whom gave me excellent feedback and suggestions.  My thanks also to Prof Ren\'{e} van der Wal for his help in the experiment mentioned in section 3.

\bibliography{nlgxairef}
\bibliographystyle{acl_natbib}

\end{document}